\PassOptionsToPackage{table}{xcolor}
\documentclass[sigconf]{aamas}  
\usepackage{booktabs}
\usepackage[utf8]{inputenc} 
\usepackage[T1]{fontenc}    
\usepackage{graphicx}
\usepackage[linesnumbered,algoruled,boxed,lined]{algorithm2e}
\usepackage{tabularx,longtable,multirow,caption}
\usepackage{graphics} 
\usepackage[style=base]{subcaption}
\usepackage{hyperref}      
\usepackage{url}            
\usepackage{amsfonts}       
\usepackage{enumitem}
\usepackage{mathrsfs}
\usepackage{amsfonts, amssymb, amsthm}
\usepackage{amsmath}

\usepackage{flushend}
\usepackage{siunitx}

\usepackage{color}
\usepackage{pgf, tikz, pgfplots}
\usetikzlibrary{shapes, arrows, automata}
\usepackage[font = small]{caption} 
\usepackage{subcaption} 

\acmDOI{}  
\acmISBN{}  
\acmConference[]{}{}{}
\acmYear{}  
\copyrightyear{}  
\acmPrice{}  

\input{mySymbol.sty}
\input{pennColors.sty}
\def\Tr{\mathsf{T}}

\begin{document}
\hyphenation{non-com-mu-ni-ca-tive}

\title{Graph Neural Networks for Decentralized Multi-Robot Path Planning}  



%

\author{Qingbiao Li$^1$, Fernando Gama$^2$, Alejandro Ribeiro$^2$, Amanda Prorok $^1$}
\thanks{$^1$Qingbiao Li and Amanda Prorok are with the Department of Computer Science and Technology, University of Cambridge, Cambridge, United Kingdom \\ (Emails: \texttt{{ql295, asp45}@cam.ac.uk})}

\thanks{$^2$Fernando Gama and Alejandro Ribeiro are with the Department of Electrical and Systems Engineering, University of Pennsylvania, Philadelphia, USA (Emails: \texttt{{fgama,aribeiro}@seas.upenn.edu})}


\begin{abstract}  
Effective communication is key to successful, decentralized, multi-robot path planning. Yet, it is far from obvious what information is crucial to the task at hand, and how and when it must be shared among robots. To side-step these issues and move beyond hand-crafted heuristics, we propose a combined model that automatically synthesizes local communication and decision-making policies for robots navigating in constrained workspaces.
Our architecture is composed of a convolutional neural network (CNN) that extracts adequate features from local observations, and a graph neural network (GNN) that communicates these features among robots. 
We train the model to imitate an expert algorithm, and use the resulting model online in decentralized planning involving only local communication and local observations. We evaluate our method in simulations {by navigating teams of robots to their destinations in 2D} cluttered workspaces. We measure the success rates and sum of costs over the planned paths. The results show a performance close to that of our expert algorithm, demonstrating the validity of our approach. In particular, we show our model's capability to generalize to previously unseen cases (involving larger environments and larger robot teams).
\end{abstract}


\maketitle



\section{Introduction} \label{sec:intro}

Efficient and collision-free navigation in multi-robot systems is fundamental to advancing mobility. The problem, generally referred to as Multi-Robot Path Planning (MRPP) or Multi-Agent Path Finding (MAPF), aims at generating collision-free paths leading robots from their origins to designated destinations. 
Current approaches can be classified as either\textit{ coupled} or \textit{decoupled}, depending on the structure of the state space that is searched. 
While coupled approaches are able to ensure the optimality and completeness of the solution, they involve \textit{centralized} components, and tend to scale poorly with the number of robots~\cite{silver2005cooperative, standley2011complete}.
Decoupled approaches, on the other hand, compute trajectories for each robot separately, and re-plan only in case of conflicts~\cite{vandenberg_Reciprocal_2008, van2005prioritized, wu_MultiRobot_2019}. This can significantly reduce the computational complexity of the planning task, but generally produces sub-optimal and incomplete solutions.
Balancing optimality and completeness with the complexity of computing a solution, however, is still an open research problem~\cite{barer2014suboptimal, sartoretti_PRIMAL_2019}. 

This work focuses on multi-robot path planning for scenarios where the robots are restricted in observation and communication range, and possess no global reference frame for localization. This naturally arises when considering physical robots equipped with hardware constraints that limit their perception and communication capabilities~\cite{matignon2012coordinated}. These scenarios impose a decentralized structure, where at any given point in time, robots have only partial information of the system state. 
In this paper, we propose a combined architecture, where we train a convolutional neural network (CNN) \cite{krizhevsky2012imagenet} that extracts adequate features from local observations, and a graph neural network (GNN) to communicate these features among robots~\cite{Gama19-Architectures} with the ultimate goal of learning a decentralized sequential action policy that yields efficient path plans for all robots. The GNN implementation seamlessly adapts to the partial information structure of the problem, since it is computed in a decentralized manner. We train this architecture to imitate an optimal coupled planner with global information that is available offline at training time. Further, we develop a dataset aggregation method that leverages an online expert to resolve hard cases, thus expediting the learning process. The resulting trained model is used online in an efficient, decentralized manner, involving communication only with nearby robots. Furthermore, we show that the model can be deployed on much larger robot teams than the ones it was trained on.




\section{Related work {and Contribution}} \label{sec:literature}
{\textbf{Related work.}}
Classical approaches to multi-robot path planning can generally be described as either centralized or decentralized. 
\emph{Centralized} approaches are facilitated by a planning unit that monitors all robots' positions and desired destinations, and returns a coordinated plan of trajectories (or way-points) for all the robots in the system. These plans are communicated to the respective robots, which use them for real-time on-board control of their navigation behavior. 
Coupled centralized approaches, which consider the joint configuration space of all involved robots, have the advantage of producing optimal and complete plans, yet tend to be computationally very expensive. 
Indeed, solving for optimality is NP-hard~\cite{yu:2013}, and although significant progress has been made towards alleviating the computational load~\cite{sharon_Conflictbased_2015, ferner_ODrM_2013}, these approaches still scale poorly in environments with a high number of potential path conflicts.

\emph{Decentralized} approaches provide an attractive alternative to centralized approaches, firstly, because they reduce the computational overhead, and secondly, because they relax the dependence on centralized units. 
This body of work considers the generation of collision-free paths for individual robots that cooperate only with immediate neighbors~\cite{desaraju_Decentralized_2012, wu_MultiRobot_2019}, or with no other robots at all~\cite{vandenberg_Reciprocal_2008}. 
In the latter case, coordination is reduced to the problem of reciprocally avoiding other robots (and obstacles), and can generally be solved without the use of communication. 
Yet, by taking purely local objectives into account, global objectives (such as path efficiency) cannot be explicitly optimized.
In the former case, it has been shown that monotonic cost reduction of global objectives can be achieved. This feat, however, relies on strong assumptions (e.g., problem convexity and invariance of communication graph \cite{wang_Separable_2018,rotkowitz_characterization_2005}) that can generally not be guaranteed in real robot systems.

\emph{Learning-based} methods have proven effective at designing robot control policies for an increasing number of tasks~\cite{rajeswaran_Generalization_2017, tobin_Domain_2017}. 
The application of learning-based methods to multi-robot motion planning has attracted particular attention due to their capability of handling high-dimensional joint state-space representations, by offloading the online computational burden to an offline learning procedure.
The work in~\cite{everett_Motion_2018} proposes a decentralized multi-agent collision avoidance algorithm based on deep reinforcement learning. Their results show that significant improvement in the quality of the path (i.e., time to reach the goal) can be achieved with respect to current benchmark algorithms (e.g., ORCA~\cite{vandenberg_Reciprocal_2008}). 
Also in recent work, Sartoretti et al.~\cite{sartoretti_PRIMAL_2019} propose a hybrid learning-based method called PRIMAL for multi-agent path-finding that uses both imitation learning (based on an expert algorithm) and multi-agent reinforcement learning.
It is note-worthy that none of the aforementioned learning-based approaches consider inter-robot communication, and thus, do not exploit the scalability benefits of fully decentralized approaches. Learning what, how, and when to communicate is key to this aim. 


Of particular interest to us is the capability of learning-based methods to handle  high-dimensional joint state-space representations, useful when planning for large-scale collective robotic systems, by offloading the online computational burden to an offline learning procedure~\cite{Tolstaya19-Flocking, khan2019graph, everett_Motion_2018}.
The fact that each robot must be able to accumulate information from other robots in its neighborhood is key to this learning procedure. From the point of view of an individual robot, its local decision-making system is incomplete, since other agents' unobservable states affect future values. \emph{The manner in which information is shared is crucial to the system's performance, yet is not well addressed by current machine learning approaches.} 
Graph Neural Networks (GNNs) promise to overcome this deficiency~\cite{scarselli_graph_2009, Gama19-Architectures}. They capture the \emph{relational} aspect of robot communication and coordination by modeling the collective robot system as a graph: each robot is a node, and edges represent communication links~\cite{prorok_Graph_2018}. Although GNNs have been applied to a number of problem domains, including molecular biology~\cite{duvenaud_Convolutional_2015}, quantum chemistry~\cite{gilmer_Neural_2017}, and simulation engines~\cite{battaglia_Interaction_2016}, they have only very recently been considered within the multi-robot domain, for applications of flocking and formation control~\cite{prorok_Graph_2018, Tolstaya19-Flocking, khan2019graph}.


{\textbf{Contributions.}}
{
The application of GNNs to the problem of multi-robot path planning is novel. Our particular GNN implementation offers an efficient architecture that operates in a localized manner, whereby information is shared over a multi-hop communication network, through {explicit communication} with nearby neighbors only~\cite{Gama19-Architectures}.
Our key contribution in this work is the development of a framework that can learn what information needs to be shared between robots, such that each robot can make a decision based on this local information; importantly, our robots only have local, relative information about their positions and goals (i.e., they possess no global reference frame).
This framework is composed of a convolutional neural network (CNN) that extracts adequate features from local observations, and a graph neural network (GNN) that learns to explicitly communicate these features among robots.
By jointly training these two components, the system is able to best determine what information is relevant for the team as a whole, and share this to facilitate efficient path planning.
The proposed model is trained to imitate a centralized coupled planner, and makes use of a novel dataset aggregation method that leverages an online expert to resolve hard cases, thus expediting the learning process.  
We achieve performance that is close to that of optimal planners in terms of success rate and flowtime (sum of path costs), while also being able to generalize to previously unseen cases, such as larger robot teams and environments.}

\section{Problem formulation} \label{sec:problem_formulatin}

Let $\ccalV = \{v_{1},\ldots,v_{N}\}$ be the set of $N$ robots. At time $t$, each robot perceives its surroundings within a given field of vision; although the robot knows where its own target destination is located, this information is clipped to the field of vision in a local reference frame (see Fig.~\ref{fig:flowchart_training}). Furthermore, we assume no global positioning of the robots. This map perceived by robot $i$ is denoted by $\bbZ_{t}^{i} \in \reals^{W_{\mathrm{FOV}} \times H_{\mathrm{FOV}}}$ where $W_{\mathrm{FOV}}$ and $H_{\mathrm{FOV}}$ are the width and height, respectively, and are determined by the field of vision radius $r_{\mathrm{FOV}}$. 

The robots can communicate with each other as determined by the communication network. We can describe this network at time $t$ by means of a graph $\ccalG_{t} = (\ccalV, \ccalE_{t}, \ccalW_{t})$ where $\ccalV$ is the set of robots, $\ccalE_{t} \subseteq \ccalV \times \ccalV$ is the set of edges and $\ccalW_{t}: \ccalE_{t} \to \reals$ is a function that assigns weights to the edges. Robots $v_{i}$ and $v_{j}$ can communicate with each other at time $t$ if $(v_{i},v_{j}) \in \ccalE_{t}$. The corresponding edge weight $\ccalW_{t}(v_{i},v_{j}) = w_{t}^{ij}$ can represent the strength of the communication (or be equal to $1$ if we are only modeling whether there is a link or not). 
{For instance, two robots $v_{i}$ and $v_{j}$, with positions $\bbp_{i}, \bbp_{j} \in \mbR^{2}$ respectively, can communicate with each other if $\|\bbp_i-\bbp_j\|\leq r_{\mathrm{COMM}}$ for a given communication radius $r_{\mathrm{COMM}} > 0$. This allows us to define an adjacency matrix $\bbS_t \in \mbR^{N \times N}$ representing the communication graph, where $[\bbS_{t}]_{ij} = s_{t}^{ij} = 0$ if $(v_{j},v_{i}) \notin \ccalE_{t}$.}

In this work, we formulate the multi-agent path planning problem as a sequential decision-making problem that each robot solves at every time instant $t$, with the objective of reaching its destination. 
More formally, the goal of this work is to learn a mapping $\ccalF$ that takes the maps $\{\bbZ_{t}^{i}\}_{v_{i} \in \ccalV}$ and the communication network $\ccalG_{t}$ at time $t$ and determines an appropriate action $\bbu_{t}$. We want the action $\bbu_{t} = \ccalF(\{\bbZ_{t}^{i}\}, \ccalG_{t})$ to be such that it contributes to the global objective of moving the robots towards their destinations in the shortest possible time while avoiding collisions with other robots and with obstacles that might be present. 
{The objective is to train the network to perform as well as a coupled centralized expert, while restricting robots to partial observations}. 
The mapping $\ccalF$ has to be restricted to involve communication only among nearby robots, as dictated by the network $\ccalG_{t}$ at each time instant $t$. 

\section{Graph Neural Networks} \label{sec:methodology}

In order to guarantee that the mapping $\ccalF$ is restricted to communications only among nearby robots, we parametrize it by means of a GNN, which is a naturally decentralized solution (Sec.~\ref{subsec:GNN}). We then train this GNN to learn appropriate actions that contribute to the global objective by means of supervised learning through an expert algorithm (i.e., imitation learning) (Sec.~\ref{subsec:imitation}).

\subsection{Graph Convolutions}

Assume that each robot has access to $F$ observations $\tbx_{t}^{i} \in \reals^{F}$ at time $t$. Let $\bbX_{t} \in \reals^{N \times F}$ be the observation matrix where each row collects these $F$ observations at each robot $\tbx_{t}^{i}$, $i=1,\ldots,N$,
\begin{equation} \label{eqn:featureMatrix}
    \bbX_{t} 
    = \begin{bmatrix}
        (\tbx_{t}^{1})^{\Tr} \\
        \vdots \\
        (\tbx_{t}^{N})^{\Tr}
      \end{bmatrix}
    = \begin{bmatrix}
        \bbx_{t}^{1} & \cdots & \bbx_{t}^{F}
      \end{bmatrix}.
\end{equation}
Note that the columns $\bbx_{t}^{f} \in \reals^{N}$ represent the collection of the observation $f$ across all nodes, for $f=1,\ldots,F$. This vector $\bbx_{t}^{f}$ is a \emph{graph signal} \cite{Ortega18-GSP}, since it assigns a scalar value to each node, $\bbx_{t}^{f}:\ccalV \to \reals$ so that $[\bbx_{t}^{f}]_{i} = x_{t}^{if} \in \reals$.

To formally describe the communication between neighboring agents, we need a concise way of describing the graph $\ccalG_{t}$ and relating it to the observations $\bbX_{t}$. {Towards this end, we use the adjacency matrix $\bbS_{t}$. We note that other matrix descriptions of the graph, such as the Laplacian matrix or the Markov matrix are possible. We generically call $\bbS_{t}$ the \emph{graph shift operator} (GSO) \cite{Ortega18-GSP}}. 


The operation $\bbS_{t} \bbX_{t}$ represents a linear combination of neighboring values of the signal due to the sparsity pattern of $\bbS_{t}$. More precisely, note that the value at node $i$ for observation $f$ after operation $\bbS_{t} \bbX_{t} \in \reals^{N \times F}$ becomes
\begin{equation} \label{eqn:graphShift}
    [\bbS_{t} \bbX_{t}]_{if} 
        = \sum_{j = 1}^{N} [\bbS_{t}]_{ij} [\bbX_{t}]_{jf}
        = \sum_{j : v_{j} \in \ccalN_{i}}
            s_{t}^{ij} x_{t}^{jf}
\end{equation}
where $\ccalN_{i} = \{v_{j} \in \ccalV : (v_{j},v_{i}) \in \ccalE_{t}\}$ is the set of nodes $v_{j}$ that are neighbors of $v_{i}$. Also, the second equality in \eqref{eqn:graphShift} holds because $s_{t}^{ij} = 0$ for all $j \notin \ccalN_{i}$.

The linear operation $\bbS_{t}\bbX_{t}$ is essentially \emph{shifting} the values of $\bbX_{t}$ through the nodes, since the application of $\bbS_{t}$ updates the value at each node by a linear combination of values in the neighborhood. With the shifting operation in place, we can define a \emph{graph convolution} \cite{Gama19-Architectures} as linear combination of shifted versions of the signal
\begin{equation} \label{eqn:graphConvolution}
    \ccalA(\bbX_{t}; \bbS_{t}) = \sum_{k=0}^{K-1} \bbS_{t}^{k} \bbX_{t} \bbA_{k}
\end{equation}
where $\{\bbA_{k}\}$ is a set of $F \times G$ matrices representing the filter coefficients combining different observations. Several noteworthy comments are in order with respect to \eqref{eqn:graphConvolution}. 
First, multiplications to the left of $\bbX_{t}$ need to respect the sparsity of the graph since these multiplications imply combinations across different nodes. Multiplications to the right, on the other hand, can be arbitrary, since they imply linear combination of observations within the same node in a weight sharing scheme. 
Second, $\bbS_{t}^{k} \bbX_{t} = \bbS_{t}(\bbS_{t}^{k-1}\bbX_{t})$ is computed by means of $k$ communication exchanges with $1$-hop neighbors, and is actually computing a summary of the information located at the $k$-hop neighborhood. Therefore, the graph convolution is an entirely local operation in the sense that its implementation is naturally distributed. 
Third, the graph convolution is actually computing the output of a bank of $FG$ filters where we take as input $F$ observations per node and combine them to output $G$ observations per node, $\ccalA(\bbX_{t};\bbS_{t}) \in \reals^{N \times G}$. There are $FG$ graph filters involved in \eqref{eqn:graphConvolution} each one consisting of $K$ filter taps, i.e., the $(f,g)$ filter can be described by filter taps $\bba^{fg} = [a_{0}^{fg},\ldots,a_{K-1}^{fg}] \in \reals^{K}$ and these filter taps are collected in the matrix $\bbA_{k}$ as $[\bbA_{k}]_{fg} = a_{k}^{fg}$.

\begin{figure*}[t]
    \centering
    \includegraphics[width=\textwidth]{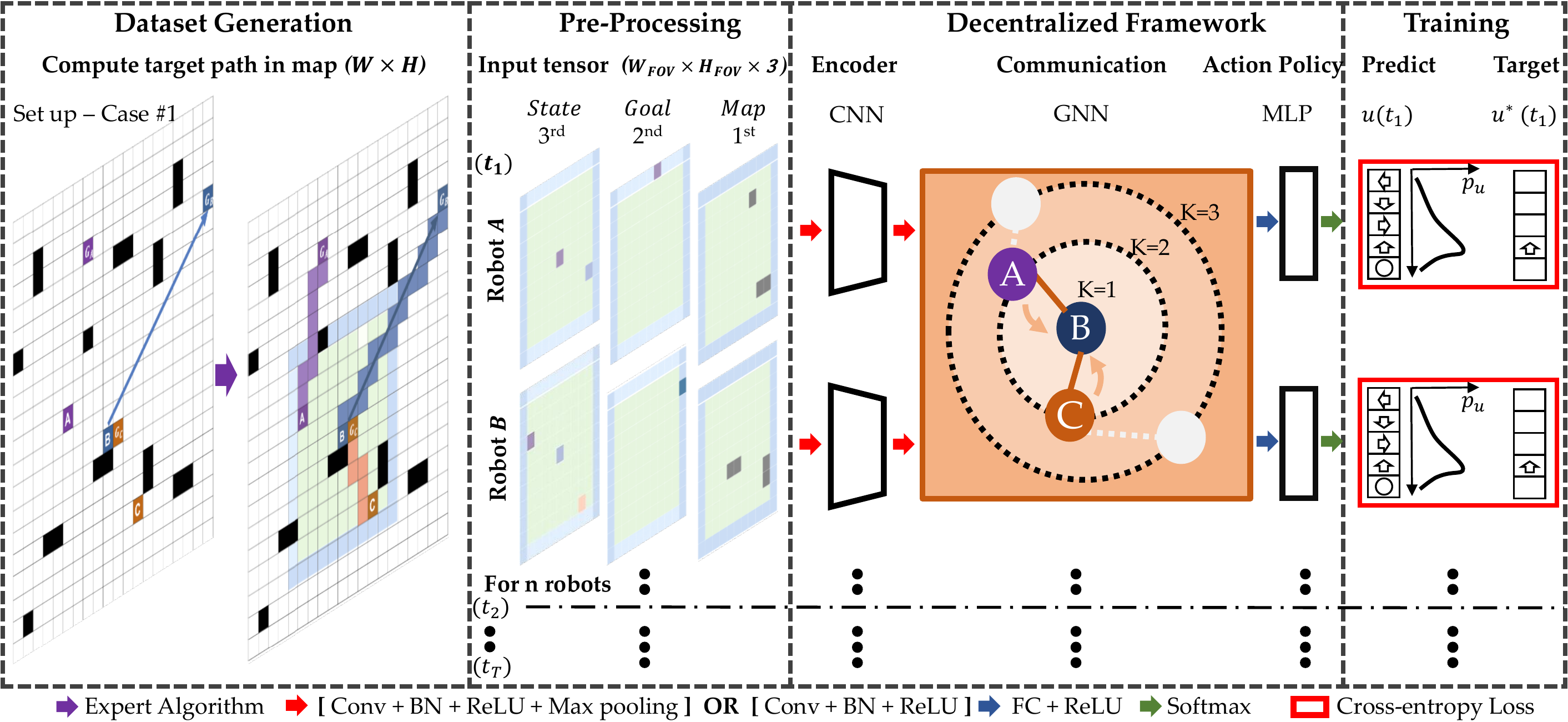}
    \caption{\normalfont Illustration of the proposed framework. \textit{(i)} The input tensor is based on a binary map representation (1st channel: partial observation of the environment; 2nd channel: the position of goal ($\bbp_{goal}^{i}$), or its projection onto the boundary of the field-of-view; 3rd channel: self (agent) at center, with other agents within its field-of-view). \textit{(ii)} The decentralized framework consists of a CNN to extract observations from the input tensor, a GNN to exchange information between the neighboring agents, and an MLP to predict the actions. \textit{(iii)} Training is performed through cross-entropy loss over a discrete action space.}
    \label{fig:flowchart_training}
    \vspace{-0.4cm}
\end{figure*}
\setlength{\belowcaptionskip}{1em}

\subsection{Graph Neural Networks} \label{subsec:GNN}
A convolutional GNN \cite{Gama19-Architectures} consists of a cascade of $L$ layers, each of which applies a graph convolution \eqref{eqn:graphConvolution} followed by a pointwise nonlinearity $\sigma: \reals \to \reals$ (also known as activation function)
\begin{equation} \label{eqn:convGNN}
    \bbX_{\ell} = \sigma \big[ \ccalA_{\ell}(\bbX_{\ell-1};\bbS) \big] \quad \text{for} \quad \ell = 1,\ldots,L
\end{equation}
where, in a slight abuse of notation, $\sigma$ is applied to each element of the matrix $\ccalA_{\ell}(\bbX_{\ell-1}; \bbS)$. The input to each layer is a graph signal consisting of $F_{\ell-1}$ observations and the output has $F_{\ell}$ observations so that $\bbX_{\ell} \in \reals^{N \times F_{\ell}}$. The input to the first layer is $\bbX_{0} = \bbX_{t}$ so that $F_{0} = F$ and the output of the last layer corresponds to the action to be taken at time $t$, $\bbX_{L} = \bbU_{t}$ which could be described by a vector of dimension $F_{L}=G$. The GSO $\bbS$ to be used in \eqref{eqn:convGNN} is the one corresponding to the communication network at time $t$, $\bbS = \bbS_{t}$. At each layer $\ell$ we have a bank of $F_{\ell}F_{\ell-1}$ filters $\ccalA_{\ell}$ described by a set of $K_{\ell}F_{\ell}F_{\ell-1}$ total filter taps $\{\bbA_{\ell k}\}_{k=0}^{K-1}$.

We note that, in the present framework, we are running one GNN \eqref{eqn:convGNN} per time instant $t$, where each time step is determined by the moment the action is taken and the communication network changes. This implies that we need to carry out $\sum_{\ell=1}^{L} (K_{\ell}-1)$ total communications before deciding on an action. Therefore, it is important to keep the GNN shallow (small $L$) and the filters short (small $K_{\ell}$).

In summary, we propose to parametrize the mapping $\ccalF$ between maps $\bbZ_{t}$ and actions $\bbU_{t}$ by using a GNN \eqref{eqn:convGNN} acting on observations $\bbX_{t} = \mathrm{CNN}(\bbZ_{t})$ obtained by applying a CNN to the input maps. We note that, by choosing this parametrization we are obtaining a mapping that is naturally distributed and that is adequately exploiting the network structure of the data.


\section{Architecture} \label{sec:framework}
The following sections describe the architecture, of which all components are illustrated in Fig.~\ref{fig:flowchart_training}.

\subsection{Processing Observations}\label{sec:CNN_preprocessing}
In an environment ($W \times H$) with static obstacles, each robot has a local field-of-view (FOV), the radius of which is defined by $r_{\mathrm{FOV}}$, beyond which it cannot `see' anything.  The robot itself is located at the center of this local observation, and does not know its global position. The data available at robot $i$ is a map $\bbZ_{t}^{i}$ of size $W_{\mathrm{FOV}} \times H_{\mathrm{FOV}}$ (Fig.~\ref{fig:flowchart_training} illustrates how we implement such partial observations).  
The input map $\bbZ_{t}^{i}$ is fed into a CNN that is run internally on each robot.  This results in a vector $\tbx_{t}^{i} \in \reals^{F}$ containing $F$ observations \eqref{eqn:featureMatrix}, $\tbx_{t}^{i} = \mathrm{CNN}(\bbZ_{t}^{i})$. {These observations can then be communicated to nearby robots. 
The intuition behind using a CNN is to process the input map $\bbZ_{t}^{i}$ into a higher-level feature tensor $\tbx_{t}^{i}$ describing the observation, goal and states of other robots. This feature tensor is then transmitted via the communication network, as described in the following section,Sec.~\ref{sec:Communication}.}



\vspace{-0.1cm}
\subsection{Communication} \label{sec:Communication}


Each individual robot communicates its compressed observation vector $\tbx_{t}^{i}$ {with neighboring robots within its communication radius $r_{\mathrm{COMM}}$} over the multi-hop communication network, whereby the number of executed hops is limited by $K$.
As described in Sec.~\ref{subsec:GNN}, we apply our GNN to aggregate and fuse the states $(\tbx_{t}^{j})$ within this $K$-hop neighborhood of robots $j \in \ccalN_{i}$, for each robot $i$. 
The output of the communication GNN is a hyper-representation of the fused information of the robot itself and its $K$-hop neighbors, which is passed to the action policy, as described in Sec.~\ref{sec:action_policy}. We note that each robot carries a local copy of the GNN, hence resulting in a localized decision-making policy.

\subsection{Action Policy}  \label{sec:action_policy}

We formulate the path-finding problem as a sequential classification problem, whereby an optimal action is chosen at each time step. 
We adopt a local multi-layer perceptron (MLP) to train our action policy network. More specifically, each node applies a MLP to the aggregated features resulting from the communication GNN. This MLP is the same across all nodes, resembling a weight-sharing scheme. The action $\tbu_{t}^{i}$ taken by robot $i$ is given by a {stochastic action policy based on} the probability distribution over motion primitives, which in our case consists of five discrete options (up, left, down, right, idle), and are represented by one-hot vectors. The final path is represented by the series of sequential actions.


\subsection{Network Architecture}
\label{sec:nn_arch}
We construct our CNN architecture by using  \texttt{Conv2d-BatchNorm2d-\\ReLU-MaxPool2d} and  \texttt{Conv2d-BatchNorm2d-ReLU} blocks sequentially three times.
All kernels are of size 3 with a stride of 1 and zero-padding. In the GNN architecture, we deploy a single layer GNN (as described in Sec.~\ref{subsec:GNN}) and set $128$ as the number of input observations $F$ and output observations $G$. Note that we can tune the filter taps $K$ for non-communication ($K=1$) and multi-hop communication ($K>1$). In the action policy, we use a linear soft-max layer to decode the output observations $G$ from the GNN with $128$ features into the five motion primitives. 



\subsection{Learning from Expert Data} \label{subsec:imitation}
To train our models, we propose a supervised learning approach based on expert data (i.e., imitation learning). We assume that, at training time, we have access to an optimal trajectory of actions $\{\bbU_{t}^{\ast}\}$ for all the robots, and the corresponding maps obtained for this trajectory $\{\bbZ_{t}^{i}\}$, collected in a training set $\ccalT = \{ (\{\bbU_{t}\},\{\bbZ_{t}^{i}\})\}$. 
Then, we train the mapping $\ccalF$ so that the output is as close as possible to the corresponding optimal action $\bbU^{\ast}$ using a cross entropy loss $\mathcal{L}(\cdot,\cdot)$. 
If the mapping $\ccalF$ is parametrized in terms of a GNN \eqref{eqn:convGNN} then this optimization problem becomes
\begin{equation} \label{eqn:trainingGNN}
    \min_{\mathrm{CNN},\{\bbA_{\ell k}\},\mathrm{MLP}} \sum_{(\{\bbU_{t}\},\{\bbZ_{t}^{i}\}) \in \ccalT} \sum_{t} {\mathcal{L}(\bbU_{t}^{\ast},\ccalF(\{\bbZ_{t}^{i}\}, \ccalG_{t}))}.
\end{equation}
We are optimizing over the filters in the CNN required to process the map as well as the set of matrices $\{\bbA_{\ell k}\}$ that contains the $\sum_{\ell=1}^{L} K_{\ell}F_{\ell-1}F_{\ell}$ learnable parameters of the communication GNN. Note that the number of parameters is independent of the size of the network $N$. 
%

Imitation learning rests on the availability of an optimal solution (further elaborated in Sec.~\ref{sec:expert_data}, below).
While this solution might be computationally expensive, or even intractable for large networks, we only need it at training time. Once trained, the GNN models can be deployed in different communication topologies~\cite{Gama19-Stability}, including those with a larger number of robots as is evidenced in the numerical experiments of Sec.~\ref{sec:results}. Given the decentralized nature of the parametrizations, the trained models are efficient in the sense that their computation is distributed among the agents, demanding only communication exchanges with one-hop neighbors.


\subsection{Expert Data Generation}
\label{sec:expert_data}
As described in our problem statement in Sec.~\ref{sec:problem_formulatin}, the robots operate in a grid world of size $W \times H$ with static obstacles randomly placed throughout the map. For each grid world, we generate \textit{cases} randomly, i.e., problem instances, which consist of pairs of start and goal positions for all robots (we also refer to this as a \textit{configuration}). We filter duplicates or invalid cases, and store the remaining cases in a setup pool, which is randomly shuffled at training time.
For each case, we generate the optimal solution. Towards this end, we run an expert algorithm: Conflict-Based Search (CBS)~\cite{sharon_Conflictbased_2015} (which is a similar approach as taken in~\cite{sartoretti_PRIMAL_2019}). This expert algorithm computes our `ground-truth paths' (the sequence of actions for individual robots), within a $300\,s$ timeout, for a given initial configuration. Our data set comprises $30{,}000$ cases for any given grid world and number of agents. This data is divided into a training set ($70\%$), a validation set ($15\%$), and a testing set ($15\%$). 


\begin{figure}[tb]
    \centering
    \includegraphics[width=\columnwidth]{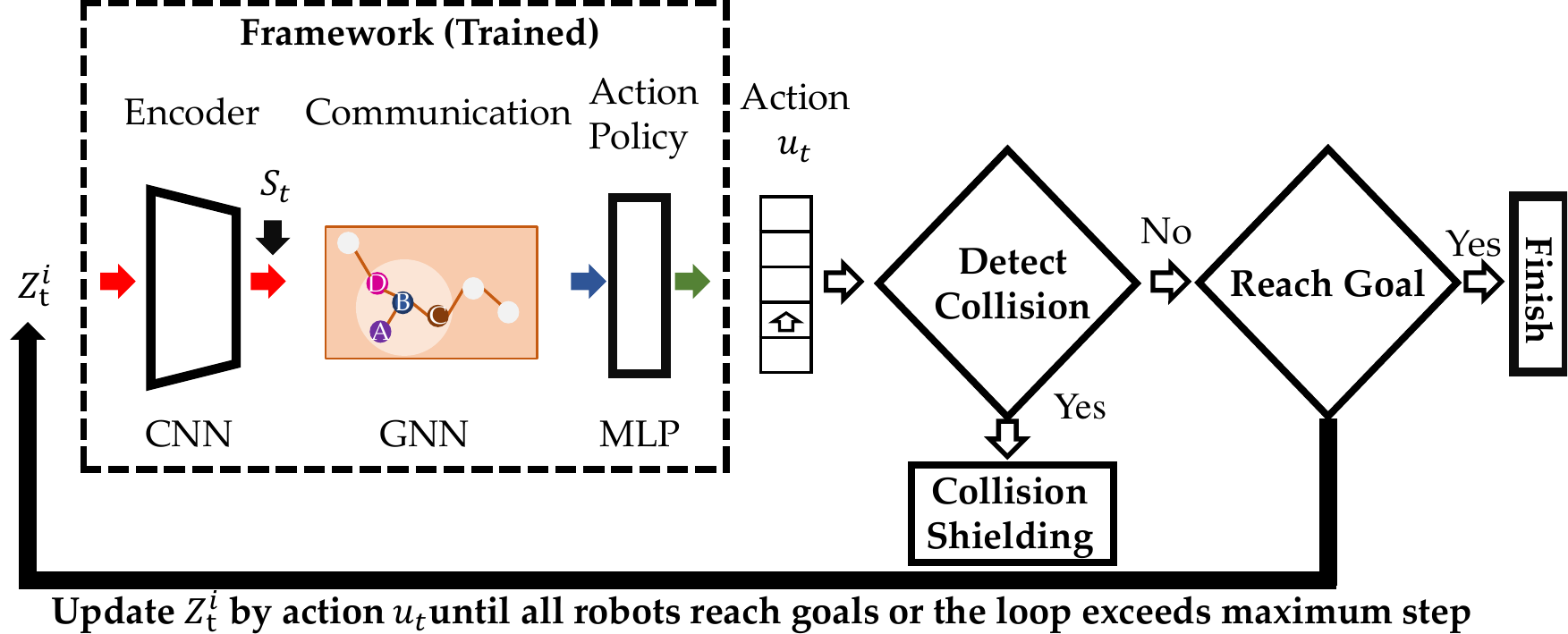}
    \caption{\normalfont Illustration of the inference stage: for each robot, the input map $\bbZ_{t}^{i}$ is fed to the trained framework to predict the action; collisions are detected and prevented by collision shielding. The input map $\bbZ_{t}^{i}$ is continuously updated until the robot reaches its goal or exceeds the timeout $T_{max}$.}
    \label{fig:flowchart_testing}
\end{figure}

\begin{algorithm}[tb]
{\small
\KwIn{Input tensor, $\bbx_{0}^{i}, i\in [0,N]$, $N$ is the number of robots; timeout $T_{max} = 3 T_{\mathrm{MP}^*}$ as explained in Sec.~\ref{sec:metrics}; Policy $\pi$}
\KwOut{Predicted paths ($\hat{\chi}_{i}$) for each robot ($i$), consisting of sequential predicted actions $\hat{u}_{t}^{i}, \forall~ t \in [0,T_{\mathrm{MP}}] $ from initial position $\bbp_{0}^{i}$}
\SetKwFunction{Inference Stage}
\SetAlgoLined
\For{$t$ in $[0, T_{max}]$}{
    \While{not all robots at their goals}{
        \For{robot $i \in \{1, \ldots, N\}$}{
        obtain input tensor $\bbx_{t}^{i}$ and adjacency matrix $\bbS_t$\;
         $\hat{u}_{t}^{i} \gets $ $\pi(\bbx_{t}^{i}, \bbS_t)$ \;
            \If{robot $i$ with action $\hat{u}_{t}^{i}$ collides with obstacle}{
                $\hat{u}_{t}^{i} \gets \mathtt{idle}$ (collision shielding)\;
                }
        }
        \eIf{robot $i$, with action $\hat{u}_{t}^{i}$, performs an edge collision with robot $j$}{
            $\hat{u}_{t}^{i} \gets \mathtt{idle}$ (collision shielding)\;}{ 
          record and update position $\bbp_{t+1}^{i}$ of robot $i$ by $\hat{u}_{t}^{i}$; {update input tensor $\bbx_{t}^{i}$ and adjacency matrix $\bbS_t$.}
          }
    }
}
Evaluate $\hat{\chi}$ according to metrics (Sec.~\ref{sec:metrics}).
\caption{Generation of sequential actions.}
\label{alg:inference_stage}
}
\end{algorithm}

\subsection{Policy Execution with Collision Shielding}
At inference stage, we execute the action policy with a protective mechanism that we name \textit{collision shielding}.
Since it is not guaranteed that robots learn collision-free paths, we require this additional mechanism to guarantee that no collisions take place. Collision shielding is implemented as follows: \textit{(i)} if the inferred action would result in a collision with another robot or obstacle, then that action is replaced by an idle action; \textit{(ii)} if the inferred actions of two robots would result in an edge collision (having them swap positions), then those actions are replaced by idle actions. It is entirely possible that robots remain stuck in an idle state until the timeout is reached. When this happens, we count it as a failure case.
The overall inference process is summarized in Alg.~\ref{alg:inference_stage} and Fig.~\ref{fig:flowchart_testing}.

\begin{algorithm}
{\small
\KwIn{Input tensor, $\bbx_{t}^{i},t\in [0,T_{\mathrm{MP}^*}], i\in [0,N]$, $N$ is the number of robots; and adjacency matrix $\bbS_t$; target actions $u_{t}^{*,i}$ generated expert algorithm; cross-entropy loss $\mathcal{L}$; learning rate $\gamma$; $(\bbx_{t}^{i},\bbS_t,u_{t}^{*,i})\in$  offline dataset $D_{\mathrm{offline}}$}
\KwOut{Proposed framework $\pi(\cdot \colon w)$}
\SetAlgoLined
$D \gets D_{\mathrm{offline}}$ \;
$\pi(\cdot \colon \bbw) \gets $ initialize parameters $\bbw$ \;
\For{$\mathrm{epoch} \in \{1, \ldots, \mathrm{epoch}_{\mathrm{max}}\}$}{
    \For{$\{\bbs_{t}^{i}, \bbS_t, u_{t}^{*,i}\}_{i = 1}^N \in D$}{
        
        \For{$i \in \{1, \ldots, N\}$}{
          $\hat{u}_{t}^{i} = \pi(\bbx_{t}^{i}, \bbS_t\colon \bbw)$ \;
          $\bbw \gets \bbw - \gamma \cdot \nabla_\bbw \mathcal{L}(\hat{u}_{t}^{i},  u_{t}^{*,i})$
        }
    }
    \If{$\mod(\mathrm{epoch}, C) = 0$}{
        \For{$n_{\mathrm{OE}}$ randomly selected cases from $D_{\mathrm{offline}}$}{
        Deploy $\pi(\cdot \colon \bbw)$ based on Alg.~\ref{alg:inference_stage}\;
        Upon timeout, deploy expert algorithm to solve failure case $D_{\mathrm{OE}}$ \;
        $D \gets D  \cup D_{\mathrm{OE}}$
        }
    }
}
\caption{{Training process with dataset aggregation.}}
\label{alg:training_stage}
}
\end{algorithm}
\setlength{\textfloatsep}{0pt}

\subsection{Dataset Aggregation during Training} \label{subsec:online_Expert}

The use of collision shielding leads to failure cases due to potential deadlocks in the actions taken by the robots, where some of them remain stuck in an idle state. To overcome such deadlocks, we propose a dataset aggregation method that makes use of an \emph{online expert} (OE) algorithm, during training. More specifically, every $C$ epochs, we select $n_{\mathrm{OE}}$ random cases from the training set and identify which ones are stuck in a deadlock situation. Then, we run the expert starting from the deadlock configuration in order to unlock them into moving towards their goal. The resulting successful trajectory is added to the training set and this extended training set is then used in the following epochs. This process is detailed in Alg.~\ref{alg:training_stage}. We note that no change is made to the validation or test sets. This dataset aggregation method is similar to the approach in DAgger~\cite{ross2011reduction}, but instead of correcting every failed trajectory, we only correct trajectories from a randomly selected pool of $n_{\mathrm{OE}}$ cases, as calls to our expert algorithm are time-consuming. Another key difference is that we need to resort to an explicit measure of failure (i.e., through the use of a timeout), since focusing on any deviations from the optimal path (as in the DAgger approach) may be misleading, because those paths may still lead to very competitive solutions in our problem setting.


\section{Performance Evaluation} 
\label{sec:evaluation}

To evaluate the performance of our method, we perform two sets of experiments, \textit{(i)} on networks trained and tested on the \textit{same} number of robots, and \textit{(ii)} on networks trained on a given number of robots, and tested on \textit{previously unseen} team sizes (both larger and smaller).

\subsection{Metrics} 
\label{sec:metrics}



    
{
1) {\textit{Success Rate ($\alpha$)}} $= n_\mathrm{success}/n$, is the proportion of successful cases over the total number of tested cases $n$. A case is considered successful (\textit{complete}) when \textit{all} robots reach their goal prior to the timeout;}

{
2) {\textit{Flowtime Increase ($\delta_{\mathrm{FT}}$)}} $=(\mathrm{FT}-\mathrm{FT}^*)/\mathrm{FT}^*$, measures the difference between the sum of the executed path lengths ($\mathrm{FT}$) and that of expert (target) path ($\mathrm{FT}^*$). We set the length of the predicted path $T^{i}=T_{\max} = 3 T_{\mathrm{MP}^*}$ (Alg.~\ref{alg:inference_stage} and Fig.~\ref{fig:flowchart_testing}), if the robot $i$ does not reach its goal.  Here, $T_{\mathrm{MP}^*}$ is the makespan of the solution generated by the expert algorithm.}

\subsection{Experimental Setup}\label{sec:exp_setup}
Our simulations were conducted using a 12-core, 3.2Ghz i7-8700 CPU and an Nvidia GTX 1080Ti GPU with 32 and 11GB of memory, respectively. The proposed network was implemented in PyTorch v1.1.0~\cite{paszke2017automatic}, and was accelerated with Cuda v10.0 APIs.  
We used the Adam optimizer with momentum $0.9$. 
The learning rate $\gamma$ scheduled to decay from $10^{-3}$ to $10^{-6}$ within 150 epochs, using cosine annealing. We set the batch size to $64$, and L2 regularization to $10^{-5}$. The online expert on the GNN is deployed every $C=4$ epochs on $n_{\mathrm{OE}}=500$ randomly selected cases from the training set. 

\subsection{Results}
\label{sec:results}
{We instantiate 600 different maps of size $20\times20$, of which 420 are used for training, 90 for validation, and 90 for testing.  We generate 50 cases for each map. The obstacle density is set to $10\%$, corresponding to the proportion of occupied over free space in the environment.}
We consider a field of view of radius $r_{\mathrm{FOV}}=4$ and a communication radius of $r_{\mathrm{COMM}} = 5$. 
At each time step, each robot runs a forwards pass of its local action policy (i.e., the trained network). At the end of each case (i.e., it is either solved or the timeout is reached), we record the length of each robot's path and the number of robots that reach their goals, to compute performance metrics according to Sec.~\ref{sec:metrics}.

\subsubsection{Effect of Communication on Flowtime and Success Rates} \label{sec:impact_K}
\begin{figure}[tb]
    \begin{subfigure}[t]{0.48\columnwidth}
        \includegraphics[width=\textwidth]{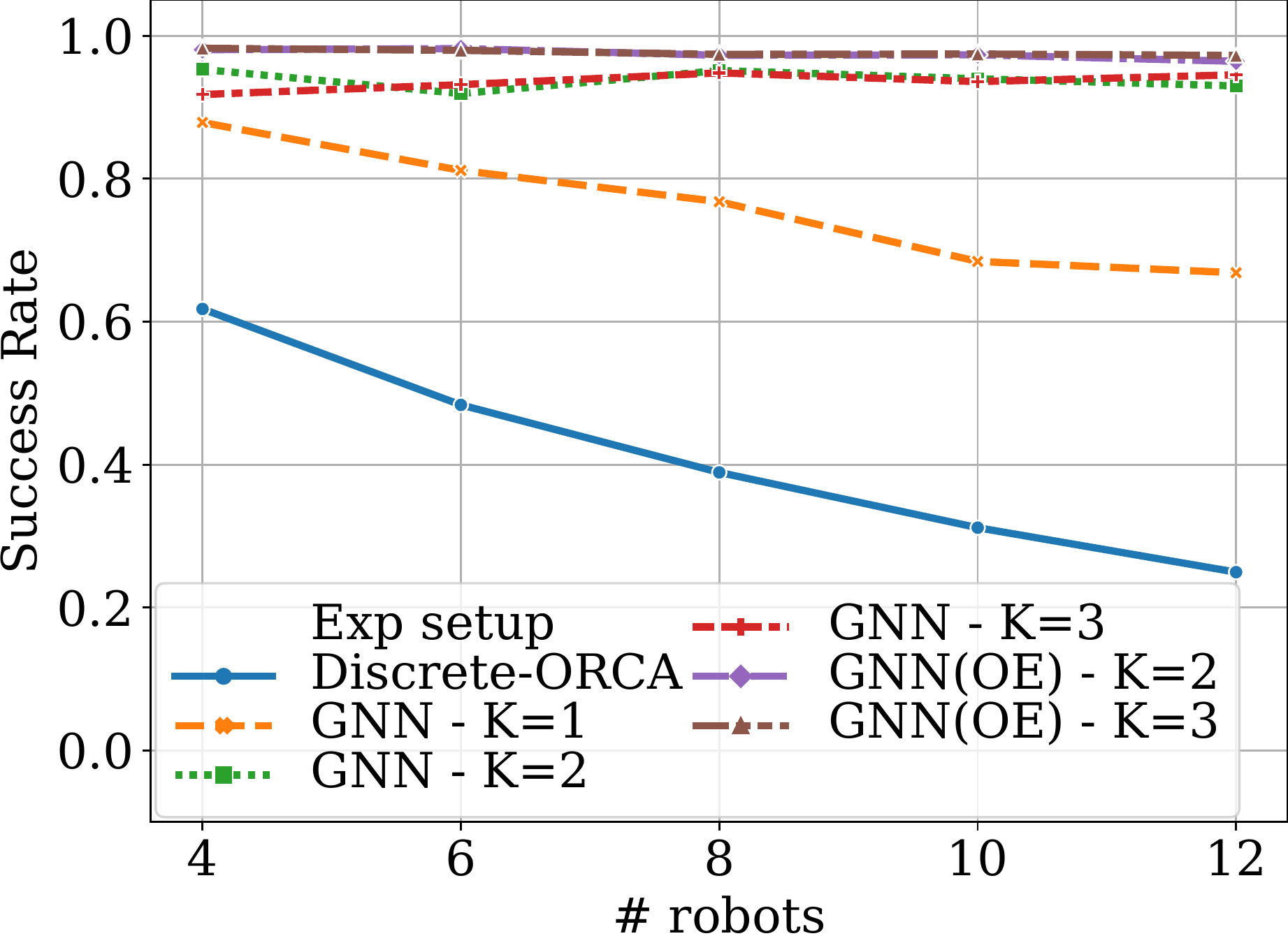}
        \caption{Success rate ($\alpha$)}
        \label{fig:results_impact_K_GNN_reachgoal}
    \end{subfigure}
        \begin{subfigure}[t]{0.48\columnwidth}
        \includegraphics[width=\textwidth]{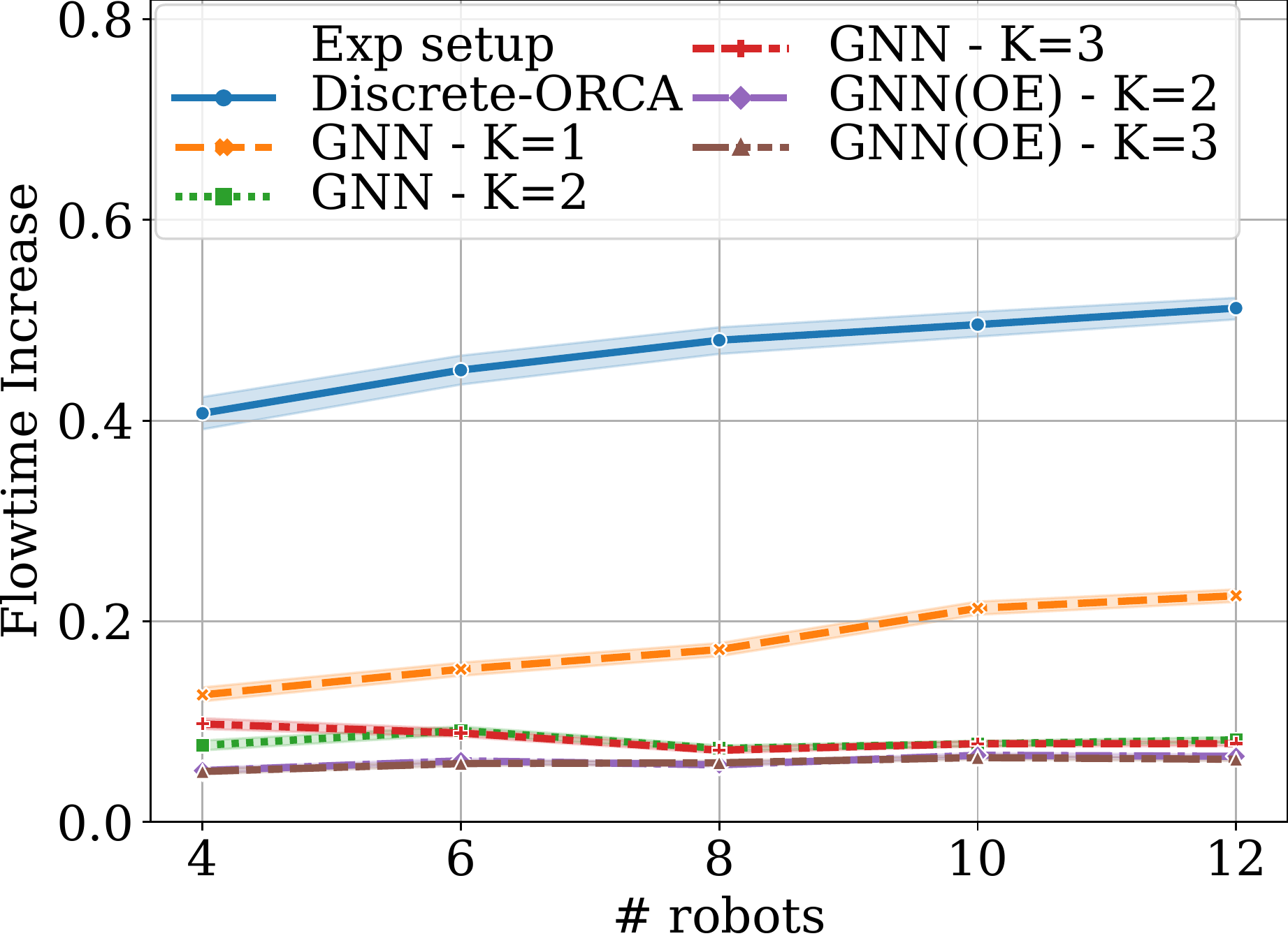}
        \caption{Flowtime increase ($\delta_{\mathrm{FT}}$)}
        \label{fig:results_impact_K_GNN_deltaFT}
    \end{subfigure}
    \caption{{{\normalfont Results for success rate ($\alpha$) and flowtime increase ($\delta_{\mathrm{FT}}$), as a function of the number of robots. For each panel, we vary the number of communication hops ($K \in [1,2,3]$), including results obtained through training with the online expert (OE). We also compare our framework with Discrete-ORCA~\cite{vandenberg_Reciprocal_2008}\cite{wang2020mobile}.}}}
    \label{fig:results_impact_K}
    \vspace{-0.3cm}
\end{figure}

Figures~\ref{fig:results_impact_K_GNN_reachgoal} and~\ref{fig:results_impact_K_GNN_deltaFT} show results for the success rate and flowtime increase, respectively, as a function of the number of robots. For each panel, we train a model for $N \in [4, 6, 8, 10, 12]$, and test it on instances of the same robot team size. In each experiment, we vary the number of communication hops ($K \in [1,2,3]$). Note that for $K=1$ there is no communication involved. {Similar to~\cite{vandenberg_Reciprocal_2008} and \cite{wang2020mobile}, we use a discrete version of a velocity-based collision-avoidance method (Discrete-ORCA) as an additional benchmark against which to test our method.}

In both figures, we see a drop in performance for larger teams, but this drop is much more pronounced for the non-communicative GNN ($K=1$). {Our framework generally outperforms the Discrete-ORCA in terms of success rate and flowtime increase.}

\subsubsection{Generalization} \label{sec:generalization}

\begin{figure}[tb]
    \centering
    \begin{subfigure}[t]{\columnwidth}
        \centering
        \includegraphics[width=\columnwidth]{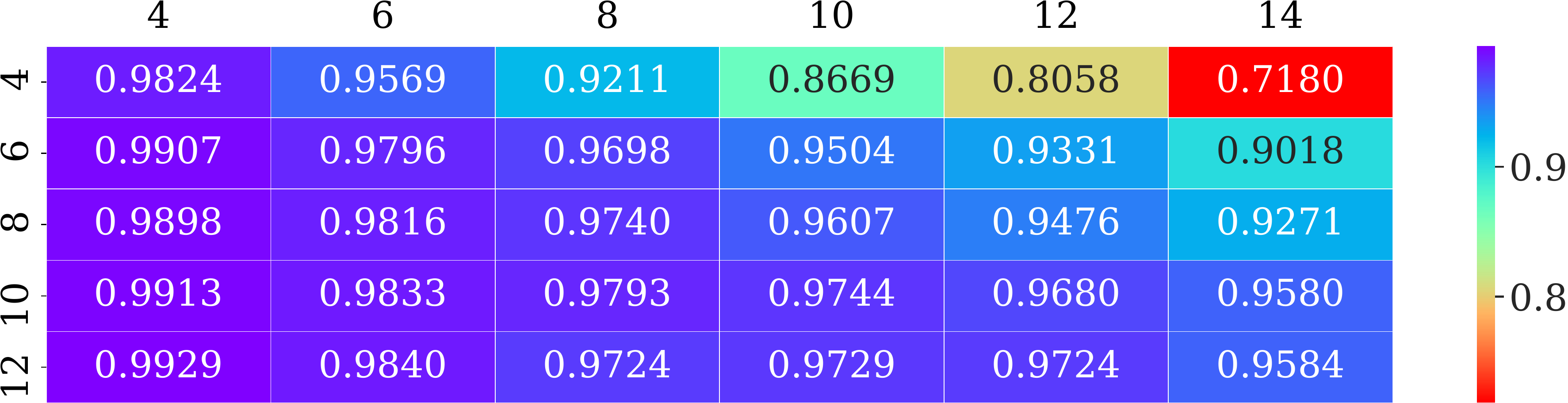}
        \caption{Success rate $\alpha$ for GNN(OE) with K=3}
        \label{fig:success_GNN_OE_K2}
    \end{subfigure}
        \begin{subfigure}[t]{\columnwidth}
        \centering
        \includegraphics[width=\columnwidth]{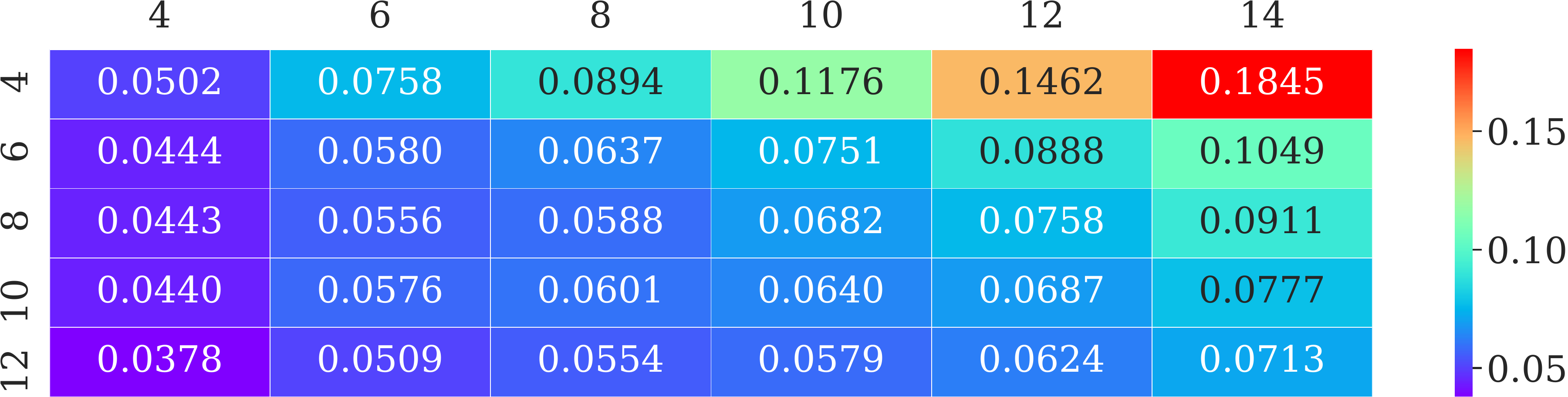}
        \caption{Flowtime increase $\delta_{\mathrm{FT}}$ for GNN(OE) with K=3}
        \label{fig:delta_FT_GNN_OE_K2}
    \end{subfigure}
     \vspace{-0.4cm}
    \caption{{\normalfont {Success rate and flowtime increase. The rows represent the number of robots on which each model was trained, and columns represent the number of robots at test time. 
    The heatmap maps performance to a color range where purple indicates the best performance and red indicates the worst performance.}}}
    \label{fig:results_generalization_heatmap}
     \vspace{-0.8cm}
\end{figure}

Fig.~\ref{fig:success_GNN_OE_K2} and~\ref{fig:delta_FT_GNN_OE_K2} summarize the generalization capability of our model for success rate and flowtime increase, respectively. The experiment was carried out by testing networks across previously unseen cases. The tables specify the number of robots \textit{trained on} in the rows, and the number of robots \textit{tested on} in the columns. The results demonstrate strong generalization capabilities.


\begin{figure}[tb]
    \begin{subfigure}[t]{0.48\columnwidth}
        \includegraphics[width=\textwidth]{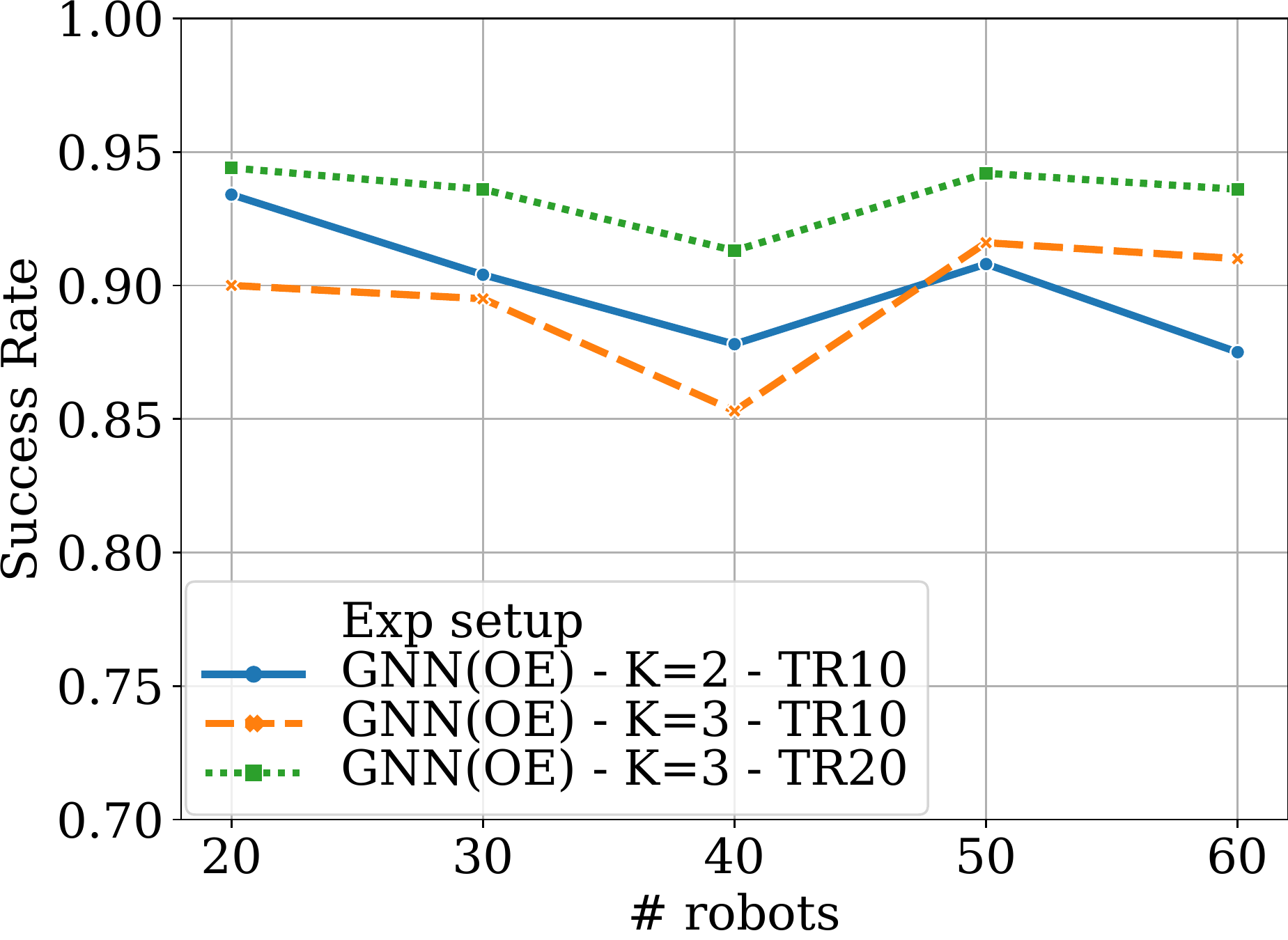}
        \caption{Success rate ($\alpha$)}
        \label{fig:results_impact_K_GNN_reachgoal_generalization}
    \end{subfigure}
        \begin{subfigure}[t]{0.48\columnwidth}
        \includegraphics[width=\textwidth]{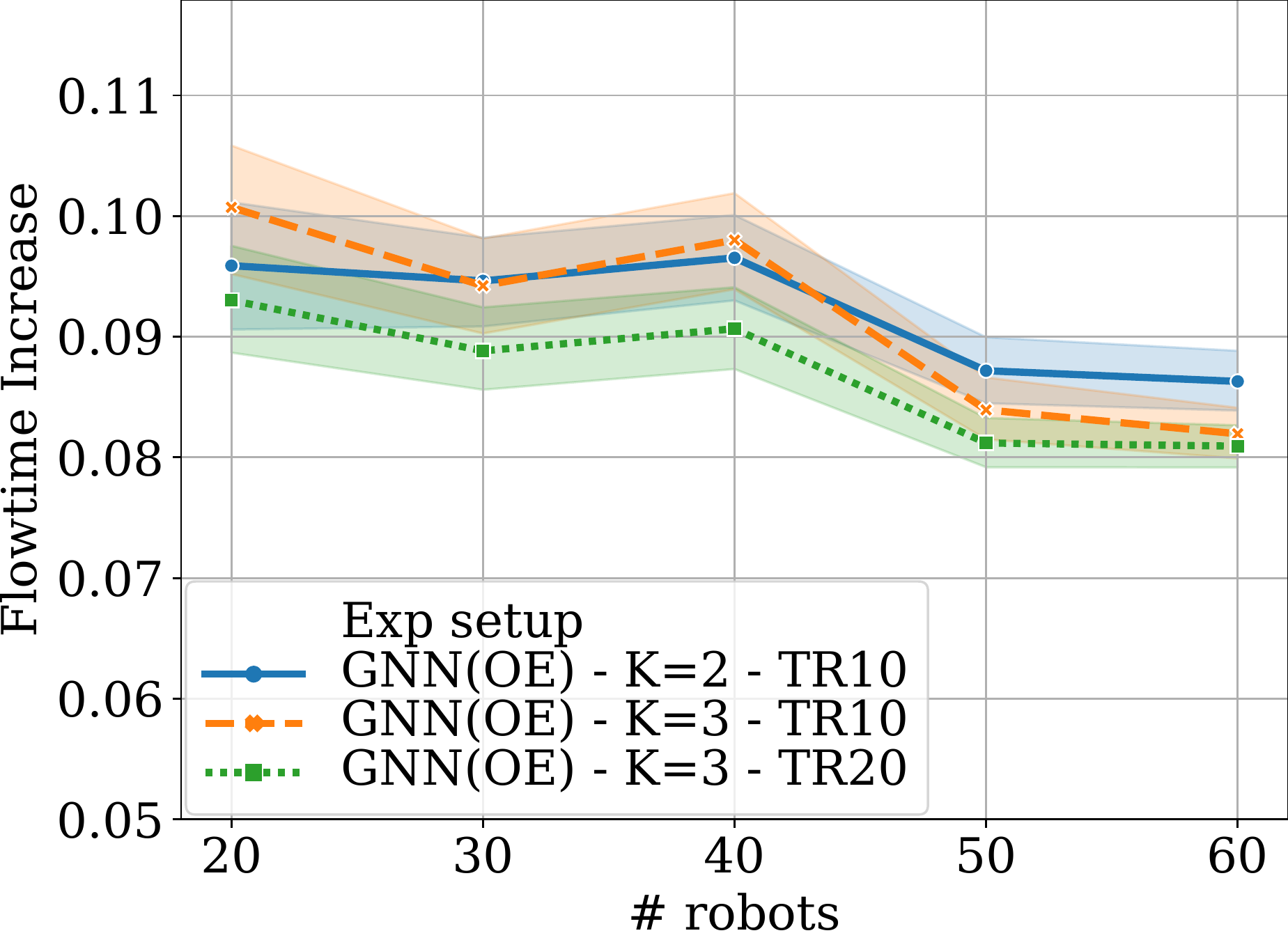}
        \caption{Flowtime increase ($\delta_{\mathrm{FT}}$)}
        \label{fig:results_impact_K_GNN_deltaFT_generalization}
    \end{subfigure}
     \vspace{-0.4cm}
    \caption{{\normalfont {Results for success rate and flowtime increase, as a function of the number robots tested on. We vary the GNN implementation ($K \in [2,3]$), trained (`TR10') on a $20\times20$ map with $10$ robots, and GNN implementation ($K = 3$) trained (`TR20') on a $28\times28$ map with $20$ robots. Testing was performed on maps that maintain constant effective robot density.}}}
    \label{fig:results_largescale_generalization}
    \vspace{-0.3cm}
\end{figure}


We perform subsequent experiments on larger robot teams to further test the generalization. 
{Results in Fig.~\ref{fig:results_largescale_generalization} show that our network, trained on only 10 robots scales to teams of sixfold size. We test the network in different grid map, where the map sizes are scaled to preserve the effective robot density. Notably, the results show \textit{no} degradation of performance}. 

We train the GNN ($K\in [2,3]$) with and without the online expert (OE) implementation on 10 robots, and {test it on 60 robots in $50\times50$ environments}, respectively. The grid maps are scaled to preserve the same effective density $\beta = \frac{n_{\text{robots}}+n_{\text{obs}}}{W\times H}$, where the number of obstacles $n_{obs}= \rho \times W \times H$, $\rho$ is the obstacle density in the map ($W \times H$) and $n_{obs}$ is the number of robots.

\begin{figure}[tb]
    \centering
    \begin{subfigure}[t]{0.48\columnwidth}
        \centering
        \includegraphics[width=\columnwidth]{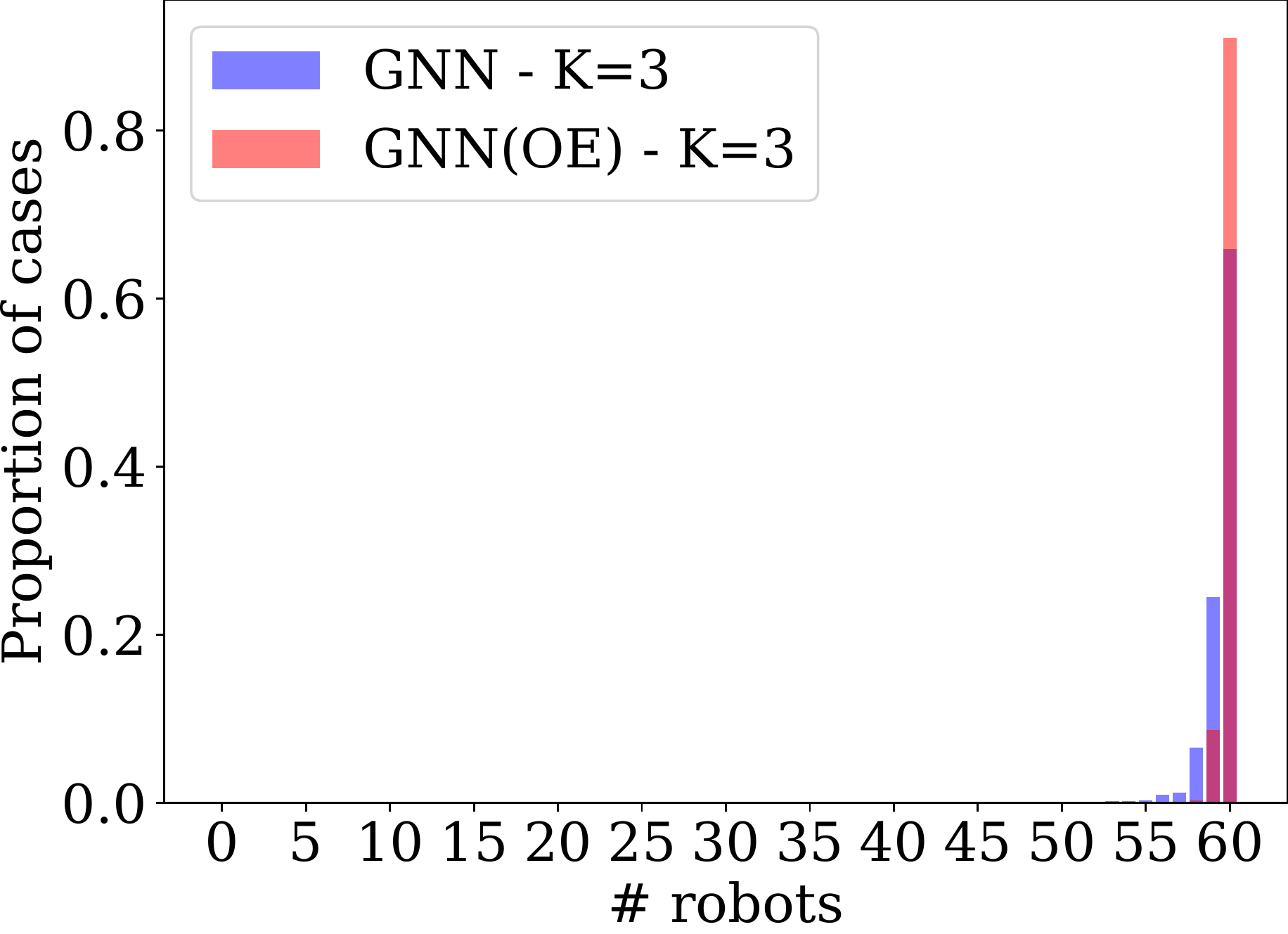}
        \caption{Effect of online expert}
        \label{fig:impact_OE}
    \end{subfigure}
        \begin{subfigure}[t]{0.48\columnwidth}
        \centering
        \includegraphics[width=\columnwidth]{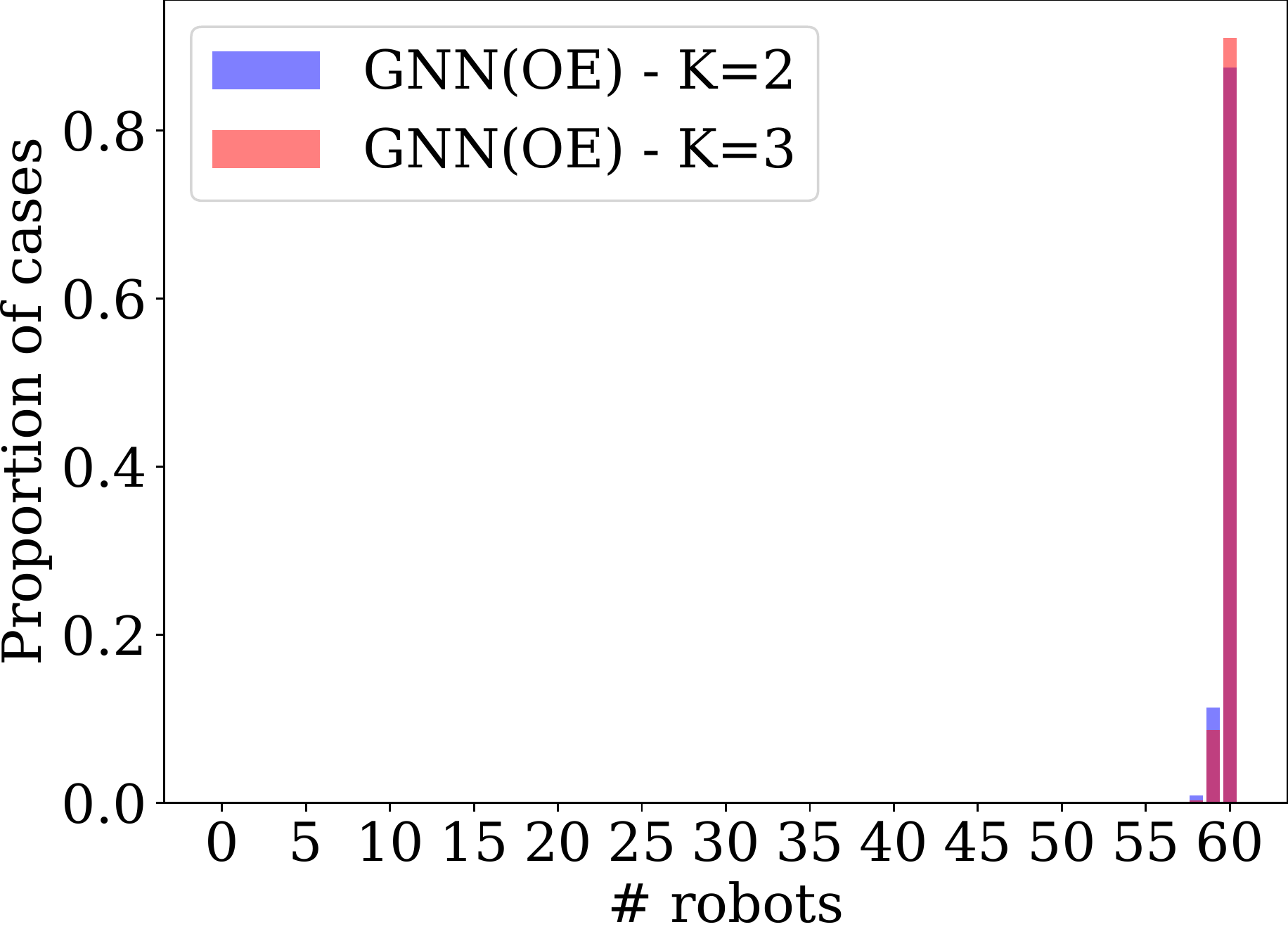}
        \caption{Effect of K}
        \label{fig:impact_K}
    \end{subfigure}
     \vspace{-0.5cm}
    \caption{ \normalfont {Histogram of proportion of cases distributed over the number of robots reaching their goal; the network with hop count $K\in [2,3]$, is trained on 10 robots and tested on 60 robots, with and without the OE.}}
    \label{fig:results_generalization}
\end{figure}

Different from our success rate metric, which only considers complete cases (all robots reach their goals), Fig.~\ref{fig:results_generalization} presents {the proportion of cases distributed over the number of robots reaching their goals}.
{The distributions show that more than 75\% of all robots \textit{always} reach their goals across all implementations. In 97\% of cases, more than 95\% of robots (57 out of 60) reach their goals. For instance, there are 995 out of 1000 cases (99.5\%), where at least 54 robots reach their goals with the GNN ($K=3$) without OE implementation (worst implementation). We see from Fig.~\ref{fig:impact_OE} how the GNN network with OE tends to generalize better than the GNN without OE, since the proportion of robots reaching the goal is larger. In Fig.~\ref{fig:impact_K}, we see how an increased communication hop count (from $K=2$ to $K=3$) contributes to a slightly larger proportion of robots reaching their goals.}

\section{Discussion and Future Work} \label{sec:Discussion}

Our results show that the decentralized framework generalizes to different numbers of robots, as seen in Sec.~\ref{sec:impact_K} and Sec.~\ref{sec:generalization}. 
We note that a single forward pass of our model (enabling a robot to predict its action) takes only $0.0019 \pm 2.15e^{-4}\,s$ on the workstation described in Sec.~\ref{sec:exp_setup}. In addition to the decentralized nature of our solution, this speed of computation is beneficial in real-world deployments, where each robot runs its own (localized) action policy. In contrast, the expert algorithm~\cite{sharon_Conflictbased_2015} is intractable for more than 14 agents in dense environments within the given timeout; this is corroborated by results in~\cite{barer2014suboptimal, sartoretti_PRIMAL_2019}.

{The experiments in in Sec.~\ref{sec:generalization} showed the capability of our decentralized policy to generalize to robot teams across different sizes.}
Fig.~\ref{fig:success_GNN_OE_K2} and Fig.~\ref{fig:delta_FT_GNN_OE_K2} showed that the framework trained in smaller robot teams ($n=4,~6$) tends to perform worse than those trained in larger teams ($n= 8,~10,~12$), across any unseen instances (larger as well as smaller in size).  The intuition for the cause of this phenomenon can be due to two main factors. Firstly, larger robot teams tend to cause more collisions, allowing the policy to learn how to plan more efficient paths more quickly. Secondly, policies trained on very small robot teams (e.g. 4 robots), tend to produce communication topologies that are idiosyncratic, and hence, may generalize more poorly. {Results in Fig.~\ref{fig:results_largescale_generalization} showed very strong generalization capabilities, with tests scaling to a factor of 6x of the instances trained on, without noticeable performance deterioration.}

We also demonstrated that the use of our online expert leads to significant improvements (as seen in Fig.~\ref{fig:results_impact_K}). 
Fig.~\ref{fig:impact_OE} shows how the GNN with the online expert was able to increase the success rate of all {60} robots reaching goal given a framework trained on 10 robots, and contribute to a right-shift of the distribution.

There are some assumptions and corresponding limitations in the current implementation, which will be improved in future work. Firstly, we assumed that communication between robots was achieved instantly without delay. Time-delayed aggregation GNNs~\cite{Tolstaya19-Flocking} can be introduced to extend our framework to handle time-delayed scenarios. 
Secondly, inter-robot live-locks and position swaps remain a challenge impeding 100\% success. 
One potential solution to this is to deploy a policy gradient to add a penalty on the action causing such scenarios. Such a strategy (e.g., as implemented in~\cite{sartoretti_PRIMAL_2019}) is harder to train, and is left for future work. 

\section{Conclusions} 
\label{sec:conclusions}
We considered the problem of collision-free navigation in multi-robot systems where the robots are restricted in observation and communication range, and possess no global reference frame for localization. We proposed a combined architecture, composed of a convolutional neural network that extracts adequate features from local observations, and a graph neural network that communicates these features among robots. The key idea behind our approach is that we jointly trained these two components, enabling the system to best determine what information is relevant for the team of robots as a whole. This approach was complemented by a data aggregation strategy that facilitated the learning process.

This work is the first to apply GNNs to the problem of multi-robot path planning. Our results show that we are very close to achieving the same performance as first-principles-based methods; in particular, we showed our model's capability to generalize to previously unseen cases involving much larger robot teams. Of particular importance is the fact that we can already scale our system to sizes that are intractable for coupled centralized solvers, while remaining computationally feasible through our decentralized approach. 

\section{Acknowledgments}
We gratefully acknowledge the support of ARL grant
DCIST CRA W911NF-17-2-0181.
A. Prorok was supported by the Engineering and Physical Sciences Research Council (grant EP/S015493/1). We gratefully acknowledge their support. We also thank Binyu Wang and Zhe Liu for their help in providing benchmark algorithms. We thank Alexandre Raymond, Hehui Zheng, Nicholas Timmons, Rupert Mitchell and Weizhe Lin for the helpful discussion. Finally, we would like to thank the anonymous reviewers for the constructive feedback and help in improving this manuscript.


\bibliographystyle{ACM-Reference-Format}  
\bibliography{AAMAS2020} 
\clearpage

\end{document}